\documentclass{article}
\usepackage{ijcai20}

\usepackage{times}
\usepackage{soul}
\usepackage{url}
\usepackage{hyperref}
\usepackage[small]{caption}
\usepackage{graphicx}
\usepackage{amsmath}
\usepackage{amssymb}
\usepackage{amsthm}
\usepackage{booktabs}
\usepackage{algorithm}
\usepackage{algorithmic}
\usepackage{color}
\usepackage{latexsym} 

\newcommand{\smallcite}[1]{{\scriptsize \cite{#1}}}

\title{Meta Cyclical Annealing Schedule:\\
A Simple Approach to Avoiding Meta-Amortization Error}

\author{
    Yusuke Hayashi$^1$
    \and
    Taiji Suzuki$^{1,2,3}$
    \affiliations
    $^1$Japan Digital Design, Japan\\
    $^2$The University of Tokyo, Japan\\
    $^3$Center for Advanced Intelligence Project, RIKEN, Japan\\
    \emails
    yusuke.hayashi@japan-d2.com,
    taiji@mist.i.u-tokyo.ac.jp
}

\begin{document}

\maketitle


\begin{abstract}
The ability to learn new concepts with small amounts of data is a crucial aspect of intelligence that has proven challenging for deep learning methods. Meta-learning for few-shot learning offers a potential solution to this problem: by learning to learn across data from many previous tasks, few-shot learning algorithms can discover the structure among tasks to enable fast learning of new tasks. However, a critical challenge in few-shot learning is task ambiguity: even when a powerful prior can be meta-learned from a large number of prior tasks, a small dataset for a new task can simply be very ambiguous to acquire a single model for that task. The Bayesian meta-learning models can naturally resolve this problem by putting a sophisticated prior distribution and let the posterior well regularized through Bayesian decision theory. However, currently known Bayesian meta-learning procedures such as VERSA suffer from the so-called {\it information preference problem}, that is, the posterior distribution is degenerated to one point and is far from the exact one.
To address this challenge, we design a novel meta-regularization objective using {\it cyclical annealing schedule} and {\it maximum mean discrepancy} (MMD) criterion. The cyclical annealing schedule is quite effective at avoiding such degenerate solutions. 
This procedure includes a difficult KL-divergence estimation, but we resolve the issue by employing MMD instead of KL-divergence.
The experimental results show that our approach substantially outperforms standard meta-learning algorithms. 
\end{abstract}

\section{Introduction}
The human visual system is efficient at grasping the main concepts of any new image from only a single or a few images. Over the last few years, few-shot learning techniques have been developed by many researchers to achieve human-level performance on image recognition tasks.
Generally, it is expected that a ``good'' few-shot learning technique should satisfy properties such as the following: (i) it is able to learn new tasks with few-shot examples efficiently, thus learning the new categories fast; (ii) the performance can be improved even as increasing numbers of input samples are given on a new task; (iii) performance on the initial tasks at training time is not sacrificed (without forgetting).

Although many few-shot classification algorithms are proposed, it is a tough task to organize the best unified framework for few-shot learning. Metric learning methods ~\cite{matchingnet,prototypical,relation-net} aim to learn a data-dependent metric to reduce intra-class distance and increase inter-class distances. Gradient-based meta-learning ~\cite{semi-amo,maml,prototypical} attempts to learn the commonalities among various tasks. MAML \cite{maml} is an effective meta-learning method that directly optimizes the gradient descent procedure for task-specific learners. In the amortized Bayesian inference framework, \cite{fewshotrecognition,amometa} proposed a method for predicting the weights of classes from activations of a pre-trained network to transfer from a high-shot classification task to a separate low-shot classification task. 
Recently,~\cite{versa} proposed a general meta-learning framework (ML-PIP) with approximate probabilistic inference and its implementation to few shot learning tasks (VERSA). ML-PIP unified a number of important approaches on meta-learning, including both gradient- and metric-based meta-learning \cite{semi-amo,maml,prototypical,fewshotrecognition,amometa} with amortized inference frameworks (neural processes) \cite{cnp,np}. It is a general framework because of the end-to-end training and supports full multi-task learning by sharing information between many tasks. In particular, VERSA replaces the optimization at test time with efficient posterior inference by generating a distribution over the task-specific parameters in a single forward pass. Therefore, this framework can amortize the cost of inference and relieve the need for second derivatives for few-shot training during test time. It is also worth noting that their inference framework is focused on the posterior predictive distribution, i.e., it aims to minimize the KL-divergence between the true and approximate predictive distributions rather than maximizing the ELBO, which is generally utilized in VAE-based methods \cite{vae}.

In the state-of-the-art models on few-shot learning tasks, amortized inference distribution is practically utilized because it is efficient and scalable to large datasets, and it requires only the specified parameters of the neural network. However, to get proper amortized inference, we need to tackle the amortization gap problem and information preference problem as stated below. As analyzed in \cite{suboptimality}, the inference mismatch between the true and approximate posterior which consists of two gaps (i) approximation gap and (ii) amortization gap. Their conclusions are that increasing the capacity of the encoder reduces the amortization error and when efficient test time inference is required, encoder generalization is important and expressive approximations in decoder are likely advantageous. Another example of the estimation difficulty of amortized inference is that cosine-similarity-based non-amortization models \cite{closerlook} achieved superior performance than those with amortization inference on few-shot learning. This implies that effective estimation methodology for amortization inference has still not been established. 

\subsection*{Our contributions in this paper are as follows:}
\begin{enumerate}
    \item We show that one of the amortization gap problems comes from the information preference problem of the latent distribution. 
    \item We adapt both the annealing method and regularization of parameter estimation in the amortized inference network to avoid the information preference problem by applying cyclical annealing schedule and maximum mean discrepancy. 
    \item Our proposal meets the ``good'' properties of few-shot learning, get better performance on standard few-shot classification tasks.
\end{enumerate}
Despite its simplicity of our proposed method, it can significantly improve the performance. Through several experimental analyses, we show that our methodology outperforms other state-of-the-art few-shot learning algorithms.

\section{Preliminaries}
\begin{figure*}[!ht]
\centering
\includegraphics[width=137mm]{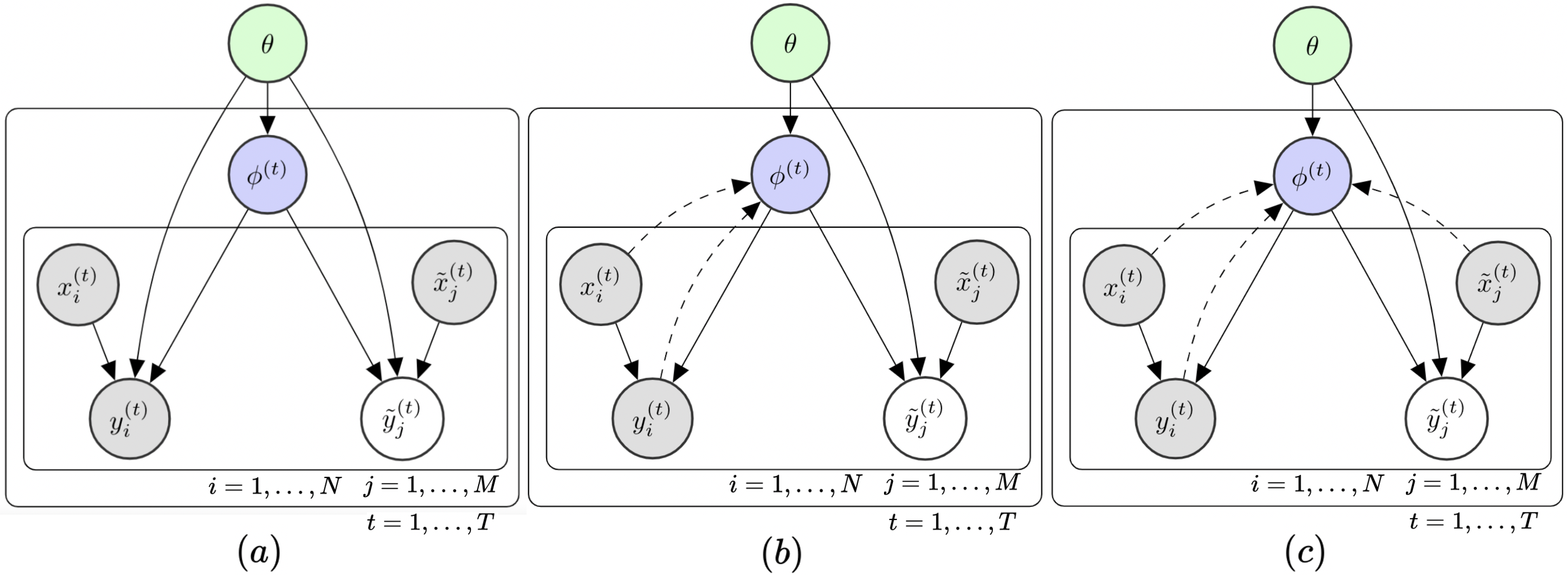}
\caption{Graphical models for meta-learning framework corresponding to our proposal method. The original graphical model ML-PIP~\protect\cite{versa} (a) is transformed into the center (b) after performing inference over $\phi^{(t)}$. The graphical model (b) represents VERSA~\protect\cite{versa} and NPs~\protect\cite{np}. We can use all observables to obtain an inference. Therefore, we can derive an additional dependency of $\phi^{(t)}$. The graphical model (c) represents the additional $\phi^{(t)}$ dependency on $\tilde{x}^{(t)}$. Dotted lines denote variational approximations. Grey node indicates ``observed''. White node indicates ``non-observed''. Purple node indicates latent variable. Green node indicates global latent variable or meta-parameter.}
\label{graphical_models}
\end{figure*}
Along with the many few-shot learning methods, a number of measures for assessing their actual performance has also being proposed. The ML-PIP model unified a number of important approaches on meta-learning, including both gradient and metric based meta-learning \cite{semi-amo,maml,prototypical,fewshotrecognition} with amortized inference framework \cite{cnp,np}. Although their method is similar to these models, it is more general, employing end-to-end training and supporting full multi-task learning by sharing information between many tasks. 
In this section, we describe the multi-task meta-learning problem that we deal with in this paper,
and we review the VERSA (implementation of ML-PIP for meta-learning) and neural processes (NPs) \cite{versa,np}. 

\subsection{Meta-learning problem}
In this paper, we mainly consider 
few-shot classification problems, in which 
we are given few-shot (say, $k$-shot) observations consisting of input-output pairs $\left\{\left(x^{c}_{i},y_{i}^{c}\right)\right\}_{i=1}^k$ for each of the $C$-classes (we call $C$-way), and we perform $C$-class classification for an unseen test input data.
We call this problem $C$-way $k$-shot meta-learning problem.
One typical approach to tackle this problem is to construct an ``encoder'' $h(x) \in \mathbb{R}^{d}$ beforehand, estimate a weight vector $W_c~(c=1,\dots,C)$ from the few-shot observations and apply the softmax operation for the linear discriminator (we call ``decoder'') $\{W_c^\top h(x)\}_{c=1}^C$. The encoder $h$ is usually trained based on other training data (which typically does not contain the $C$-class few-shot observations) so that $h$ extracts informative features that can distinguish the unseen classes. 
For the training phase, we are given training data for several tasks, $\mathcal{D}^{\left( t \right)}_{\mathrm{tr}} = \mathcal{D}^{\left( t \right)}_{S,\mathrm{tr}} \cup \mathcal{D}^{\left( t \right)}_{Q,\mathrm{tr}}$ ($t$ is the task index: $t=1,\dots,T$) where 
$\mathcal{D}^{(t)}_{S,\mathrm{tr}} = \left\{\left(x_{i}^{(t)}, y_{i}^{(t)}\right)\right\}_{i=1}^{N}$ and $\mathcal{D}^{(t)}_{Q,\mathrm{tr}} =  \left\{\left(\tilde{x}^{(t)}_{j},\tilde{y}^{(t)}_{j}\right)\right\}_{j=1}^{M}$, the number of observations $N=k \times C$ for each task is supposed to be small. Based on the support dataset $\mathcal{D}^{(t)}_{S,\mathrm{tr}}$, we train the encoder $h$ and the network which produces the weight vector $\{W_c\}_{c=1}^C$. This procedure can be seen as a kind of learning a training procedure. In the test phase, we are given test data $\mathcal{D}^{\left( t \right)}_{\mathrm{ts}} = \mathcal{D}^{\left( t \right)}_{S,\mathrm{ts}} \cup \mathcal{D}^{\left( t \right)}_{Q,\mathrm{ts}}$ ($t$ is the task index: $t=1,\dots,T^{\prime}$) where 
$\mathcal{D}^{(t)}_{S,\mathrm{ts}} = \left\{\left(x_{i}^{(t)}, y_{i}^{(t)}\right)\right\}_{i=1}^{N}$ and $\mathcal{D}^{(t)}_{Q,\mathrm{ts}} =  \left\{\left(\tilde{x}^{(t)}_{j},\tilde{y}^{(t)}_{j}\right)\right\}_{j=1}^{M}$ of new unseen tasks. Based on the support dataset $\mathcal{D}^{(t)}_{S,\mathrm{ts}}$, the encoder produces the new weight vector $\{W_c\}_{c=1}^C$. In the few-shot learning setting, $\tilde{y}_{j}^{(t)}$ is a class label among $C$-classes and the total number of data is $M$.

\subsection{Meta-Learning via amortized Bayesian inference}
VERSA is a Bayesian meta-learning framework and is also used in the few-shot classification task. Its graphical model is shown in Figure \ref{graphical_models} (a). VERSA consists of two parts: an {\it encoder} and an {\it amortization network}. The encoder $h(x) = h_\theta(x)$ maps an input to a feature vector,
and its parameter is denoted by $\theta$ (we call global latent variable). We use the same $\theta$ across all the tasks. As we have described in the previous section, this encoder is trained through the support dataset $\mathcal{D}^{(t)}_{S,\mathrm{tr}}$.~When a new task appears at test time, the same encoder as the one estimated at the training time is used for test as well; that is, for a newly observed task $t$, 
shared statistical encoder $h$ is fed $\tilde{x}^{(t)}$ as input and it outputs $h_{\theta}(\tilde{x}^{(t)})$ as a representation of the input $\tilde{x}$. The amortization network outputs the predictive distribution from the representation of the input $h_\theta(\tilde{x}^{(t)})$. It is characterized by the task specific parameter $\lambda^{(t)}$ which represents a network that maps the encoded input $h_{\theta}(\tilde{x})$ to the parameters of the approximated posterior distribution of the parameters of the output label $\tilde{y}^{(t)}$. In VERSA model, $\lambda^{(t)}$ have to be trained with few-shot samples at training time using the training data $\mathcal{D}^{(t)}_{\mathrm{tr}}$. In practice, as the amortized function, essentially a neural network, is estimated to take a representation variable as input, and outputs the mean and variance parameter for predictive distribution of each task. 

For the few-shot classification task, VERSA encodes the class $c \in \{1,\dots,C\}$ by the average of the encoded-input $h_\theta(x_i^c)$: $\bar{h}_c = \sum_{i=1}^k h_\theta(x_i^c)$.~This acts like the weight vector $W_c$ for the classification. Basically, the predictive distribution for $\tilde{y}$ is given by the softmax value of $\{\bar{h}_c^\top h_\theta(\tilde{x})\}_{c=1}^C$. To obtain the approximated posterior predictive distribution, VERSA generates $\phi_{c,l}$
as a stochastic version of $\bar{h}_c$ from the Gaussian distribution with mean $\phi_\mu$ and variance $\phi^2_\sigma$ specified by the output of $\lambda^{(t)}$ $\left( \phi_{c.l} \sim \mathcal{N}(\phi_{\mu}(\bar{h}_c)),\phi^{2}_{\sigma}(\bar{h}_{c})) \right)$, and sample the predictive distribution corresponding to $\frac{1}{L} \sum_{l=1}^L \textup{softmax}\left(\left\{\phi_{c,l}^{\top} h_{\theta}(\tilde{x})\right\}^{C}_{c=1}\right)$.

This framework approximates the posterior predictive distribution by an amortized distribution as follows. Here, the predictive distribution of the test output $\tilde{y}$ given the input $\tilde{x}$ and the few-shot sample $h_{\theta}(x)$ is given as 
\begin{eqnarray}
  & \displaystyle \hspace{-58pt} p(\tilde{y}|\tilde{x}, h_{\theta}(x), \theta) = \int p(\tilde{y}, \phi | \tilde{x}, h_{\theta}(x), \theta)d\phi \nonumber \\
  & \displaystyle \hspace{50pt} = \int p(\tilde{y} | \tilde{x}, \phi, \theta) p(\phi | \tilde{x}, h_{\theta}(x), \theta)d\phi. 
\end{eqnarray}
where $p(\tilde{y} | \tilde{x}, \phi, \theta)$ corresponds to softmax function. However, the posterior distribution of $\phi$ is difficult to calculate. Therefore, VERSA approximates the predictive distribution by the amortized distribution $p(\phi | \tilde{x}, h_{\theta}(x), \theta)$ by utilizing the approximated posterior distribution $q_{\lambda}(\phi | h_{\theta}(x), \theta)$.
VERSA employs a Gaussian distribution as the approximated posterior $q_{\lambda}(\phi | h_{\theta}(x), \theta)$ which is characterized by the network output: $\phi \sim \mathcal{N}(\phi_{\mu}(h_{\theta}(x)),\phi^{2}_{\sigma}(h_{\theta}(x)))$. Then, the amortized predictive distribution is given as
\begin{eqnarray}
  & \displaystyle \hspace{-45pt} q_{\lambda}(\tilde{y}|\tilde{x}, h_{\theta}(x), \theta) = \int q_{\lambda}(\tilde{y}, \phi | \tilde{x}, h_{\theta}(x), \theta)d\phi \nonumber \\
  & \displaystyle \hspace{55pt} = \int p(\tilde{y} | \tilde{x}, \phi, \theta) q_{\lambda}(\phi | h_{\theta}(x), \theta)d\phi.
\end{eqnarray}
\begin{figure*}[!ht]
\centering
\includegraphics[width=163mm]{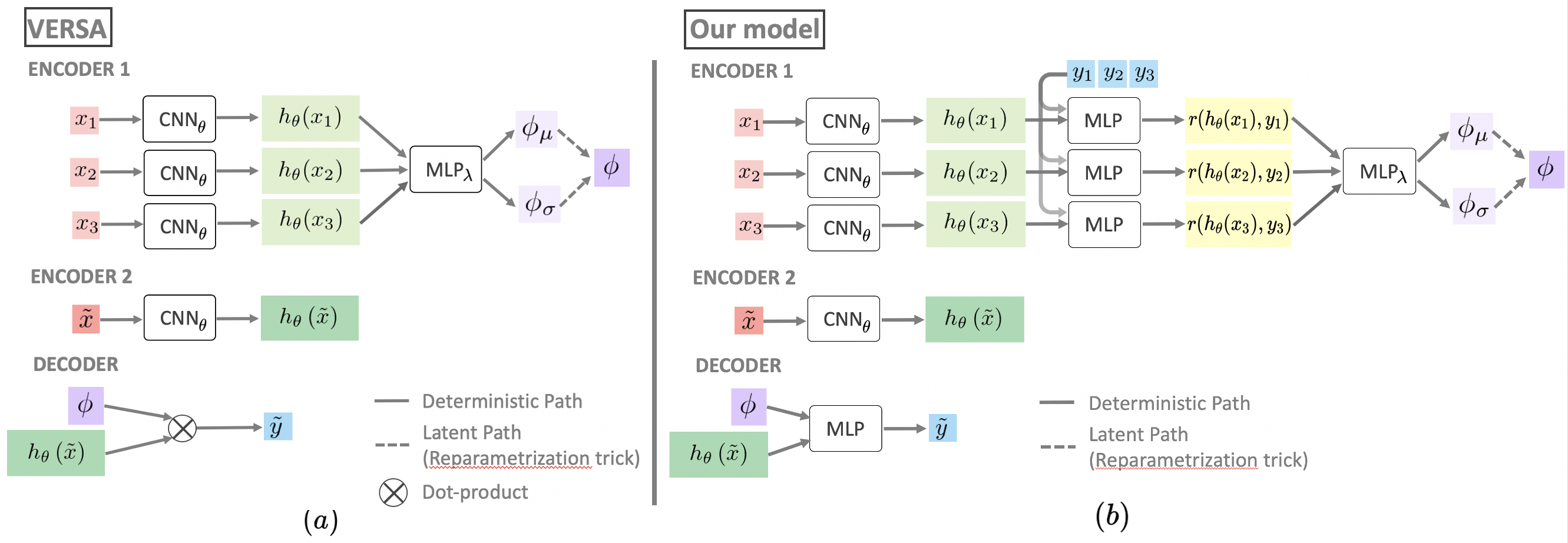}
\caption{This figure describes implementation differences between VERSA and our model. We replace encoder1 for a more precise implementation with the graphical model (b) [see Figure~\protect\ref{graphical_models}]. Then, representations of each (x) replace representations of each (x, y) pair. This approach is in a similar manner to NPs~\protect\cite{cnp,np}.}
\label{our_model}
\end{figure*}
Since VERSA wants to approximate the predictive distribution as accurate as possible, the end-to-end stochastic training objective to be minimized for $\theta$ and $\lambda = \{\lambda^{(t)}\}_{t=1}^T$ is given as follows:
\begin{eqnarray}
  & \displaystyle \hspace{-55pt} \mathcal{L}(\theta, \lambda) = -\mathbb{E}_{p(\tilde{y}, \tilde{x}, h_{\theta}(x))}\left[\mathrm{log} \ q_{\lambda}(\tilde{y} | \tilde{x}, h_{\theta}(x), \theta)\right] \nonumber \\
  & \displaystyle \hspace{2pt} = -\mathbb{E}_{p(\tilde{y},\tilde{x}, h_{\theta}(x))}\left[\mathrm{log} \int p(\tilde{y} | \tilde{x}, \phi, \theta)\ q_{\lambda}(\phi | h_{\theta}(x), \theta) d\phi \right] \nonumber \\
  & \displaystyle \hspace{-15pt} \simeq -\frac{1}{MT}\sum^{M}_{j}\sum^{T}_{t} \mathrm{log} \left( \frac{1}{L} \sum^{L}_{l} p(\tilde{y}^{(t)}_{j} | \tilde{x}^{(t)}_{j}, \phi^{(t)}_{l}, \theta) \right).
\label{eq:Lhatthetaphi}
\end{eqnarray}
However, in general, learning ``good'' latent code is difficult because even when a powerful prior can be meta-learned from a large number of prior tasks, a small dataset for a new task can simply be too ambiguous to acquire a single accurate model. Here, we consider a more general objective which includes the regularization term:
\begin{eqnarray}
  & \displaystyle \hspace{-20pt} \mathbb{D}_{\rm{KL}}\left[p(\tilde{y},\phi|\tilde{x}, h_{\theta}(x), \theta) || q_{\lambda}(\tilde{y},\phi|\tilde{x}, h_{\theta}(x), \theta)\right] \nonumber \\
  & \displaystyle \hspace{-0pt} = \mathbb{D}_{\rm{KL}}\left[p(\tilde{y}|\tilde{x}, h_{\theta}(x), \theta) || q_{\lambda}(\tilde{y}|\tilde{x}, h_{\theta}(x), \theta)\right] \nonumber \\
  & \displaystyle \hspace{10pt} + \mathbb{D}_{\rm{KL}}\left[p({\phi}| \tilde{x}, h_{\theta}(x), \theta) || q_{\lambda}({\phi}|h_{\theta}(x),\theta)\right].
\label{eq:generalizedL}
\end{eqnarray}
In the objective, the KL-divergence $\mathbb{D}_{\rm{KL}}\left[p({\phi}| \tilde{x}, h_{\theta}(x), \theta) || \right.$ $\left. q_{\lambda}({\phi}|h_{\theta}(x),\theta)\right]$ between the posterior distributions work as regularization. Unfortunately, the conditional prior
$p({\phi}| \tilde{x}, h_{\theta}(x), \theta)$ in the above expression is intractable. To resolve this issue, we instead use an approximated posterior $q_{\lambda}({\phi}| \tilde{x}, h_{\theta}(x),\theta)$, which gives:
\begin{eqnarray}
  & \displaystyle \hspace{-20pt} \mathbb{D}_{\rm{KL}}\left[p(\tilde{y},\phi|\tilde{x}, h_{\theta}(x), \theta) || q_{\lambda}(\tilde{y},\phi|\tilde{x}, h_{\theta}(x), \theta)\right] \nonumber \\
  & \displaystyle \hspace{-0pt} \simeq \mathbb{D}_{\rm{KL}}\left[p(\tilde{y}|\tilde{x}, h_{\theta}(x), \theta) || q_{\lambda}(\tilde{y}|\tilde{x}, h_{\theta}(x), \theta)\right] \nonumber \\
  & \displaystyle \hspace{10pt} + \mathbb{D}_{\rm{KL}}\left[q_{\lambda}({\phi}|\tilde{x},h_{\theta}(x),\theta) || q_{\lambda}({\phi}|h_{\theta}(x),\theta)\right].
\label{eq:approximatedL}
\end{eqnarray}

It is interesting to note that replacing the feature vector $h_{\theta}(x)$ with $r(h_{\theta}(x),y)$, we got the NPs objective from the above objective:
\begin{eqnarray}
  & \displaystyle \hspace{-12pt} \mathbb{D}_{\rm{KL}}\left[p(\tilde{y},\phi|\tilde{x}, r(h_{\theta}(x),y), \theta) || q_{\lambda}(\tilde{y},\phi|\tilde{x}, r(h_{\theta}(x),y), \theta)\right] \nonumber \\
  & \displaystyle \hspace{-2pt} \simeq \mathbb{D}_{\rm{KL}}\left[p(\tilde{y}|\tilde{x}, r(h_{\theta}(x),y), \theta) || q_{\lambda}(\tilde{y}|\tilde{x}, r(h_{\theta}(x),y), \theta)\right] \nonumber \\
  & \displaystyle \hspace{7pt} + \mathbb{D}_{\rm{KL}}\left[q_{\lambda}({\phi}|\tilde{x},r(h_{\theta}(x),y),\theta) || q_{\lambda}({\phi}|r(h_{\theta}(x),y),\theta)\right].
\label{eq:neuralprocessesL}
\end{eqnarray}
where the function $r$ is the neural network. NPs combines the strengths of neural networks and Gaussian processes to achieve both flexible learning and fast prediction in stochastic processes. While VERSA uses linear discriminator as a decoder, NPs uses neural network as it. 
Both models are represented in Figure \ref{graphical_models} (b).

The point is that, in VERSA and NPs, the central (stochastic) function being learnt has a form $\tilde{y} = f(\tilde{x}, \mathcal{D}_{S}, \theta)$, of an output $\tilde{y}$ given an input $\tilde{x}$, a support dataset $\mathcal{D}_{S}$ and the encoder's parameter (global latent variable) $\theta$. 

\subsection{Information preference property}
\label{sec:InfPreProp}
As in VERSA and NPs, we consider the following generative process for $\tilde{y}$,
\begin{eqnarray}
  & \displaystyle \phi \sim p(\phi| \tilde{x}, \mathcal{D}_{S}, \theta), \hspace{10pt} \tilde{y} \sim p(\tilde{y}| \tilde{x}, \phi, \theta).
\end{eqnarray}
where $p(\phi| \tilde{x}, \mathcal{D}_{S}, \theta)$ is the prior and $p(\tilde{y}|\tilde{x},\phi,\theta)$ is given by a generative model with parameter $\theta$. Under ideal conditions, optimizing the objective using sufficiently flexible model families for $p(\tilde{y}|\tilde{x},\phi,\theta)$ and $q_{\lambda}(\phi | \mathcal{D}_{S}, \theta)$ over $\theta, \lambda$ will achieve both goals of correctly capturing $p(\tilde{y}, \tilde{x}, \mathcal{D}_{S}, \theta)$ and performing correct amortized inference. However, this approach suffers from the following problem: the decoder tends to neglect the latent variables $\phi$ altogether, that is, the mutual information between $\phi$ and $\tilde{y}$ conditioned on $(\tilde{x}, \theta)$ becomes negligibly small. For example,
\begin{eqnarray*}
& \displaystyle \hspace{-0pt} \frac{p(\phi | \tilde{x}, \mathcal{D}_{S}, \theta)}{q_{\lambda}(\phi | \mathcal{D}_{S}, \theta)} = \frac{p(\phi, \tilde{x} | \mathcal{D}_{S}, \theta)}{p(\phi | \mathcal{D}_{S}, \theta) p(\tilde{x} | \mathcal{D}_{S}, \theta)} \cdot \frac{p(\phi | \mathcal{D}_{S}, \theta)}{q_{\lambda}(\phi | \mathcal{D}_{S}, \theta)}.
\end{eqnarray*}
Therefore, 
\begin{eqnarray*}
& \displaystyle \hspace{-50pt} \mathbb{E}_{p(\tilde{x} | \mathcal{D}_{S}, \theta)}\left[\mathbb{D}_{\mathrm{KL}}\left[p(\phi | \tilde{x}, \mathcal{D}_{S}, \theta) \| q_{\lambda}(\phi | \mathcal{D}_{S}, \theta)\right]\right] \nonumber \\
& \displaystyle \hspace{-18pt} =\int p(\phi, \tilde{x} | \mathcal{D}_{S}, \theta) \log \frac{p(\phi | \tilde{x}, \mathcal{D}_{S}, \theta)}{q_{\lambda}(\phi | \mathcal{D}_{S}, \theta)} d \phi d \tilde{x} \nonumber \\
& \displaystyle \hspace{20pt} =\mathbb{I}_{p}\left[\phi, \tilde{x} | \mathcal{D}_{S}, \theta\right]+\mathbb{D}_{\mathrm{KL}}\left[p(\phi | \mathcal{D}_{S}, \theta) \| q_{\lambda}(\phi | \mathcal{D}_{S}, \theta)\right].
\label{eq:mutualinfo}
\end{eqnarray*}
The above objective tells us that the more the learning procedure proceeds (that is, the left-hand side reduces), the more the mutual information between $\phi$ and $\tilde{x}$ decreases. It follows that the mutual information between $\phi$ and $\tilde{y}$ decrease. Intuitively, the reason is because the distribution of $\phi$ tends to shrink to a single point and $\phi$ is almost uniquely identified by a given $\tilde{x}$ and $\theta$. This is undesirable because such a shrunken posterior of $\phi$ is far from the true posterior and severely lose variation of the posterior sampling for $\phi$. This effect, which we shall refer to as the information preference problem, was studied in the VAE framework with a coding efficiency argument \cite{vae}. In the VAE framework, the issue causes two undesirable outcomes: (1) the learned features are almost identical to the uninformative Gaussian prior for all observed tasks; and (2) the decoder completely ignores the latent code, and the learned model reduces to a simpler model \cite{cyclical}.

\section{Proposed Method}
As seen in the previous section, a unified Bayesian inference framework ML-PIP and its implementation VERSA and NPs utilize the amortized inference distribution because it is efficient and scalable to large datasets, and it requires only the specified parameters of the neural network. However, as we have seen in the previous section, the regularization in the objective causes the information preference problem (which is a well-known issue in the VAE framework) and inaccurate estimation of amortized inference distributions.

One approach to remedy this issue is to introduce a hyperparameter $\beta$ to control the strength of regularization \cite{betavae}. Furthermore, \cite{cyclical} found that scheduling $\beta$ during the model training highly improve the performance. In addition, \cite{infovae} reported an alternative approach: replacing the KL-divergence of the latent distributions in the objective with the alternative divergence.

However, these previous studies applied their methodologies to only single task learning framework. In contrast, this paper considers the {\it cyclical annealing schedule} for $\beta$ during multi-task learning (meta-training) and replacing the divergence with the {\it maximum mean discrepancy} criterion. This procedure leads high mutual information between $\tilde{x}$ and $\phi$, which encourages the model to use the latent code and avoids the information preference problem. 

\subsection{Cyclical annealing schedule}
\begin{figure}[h]
\centering
\includegraphics[width=83mm]{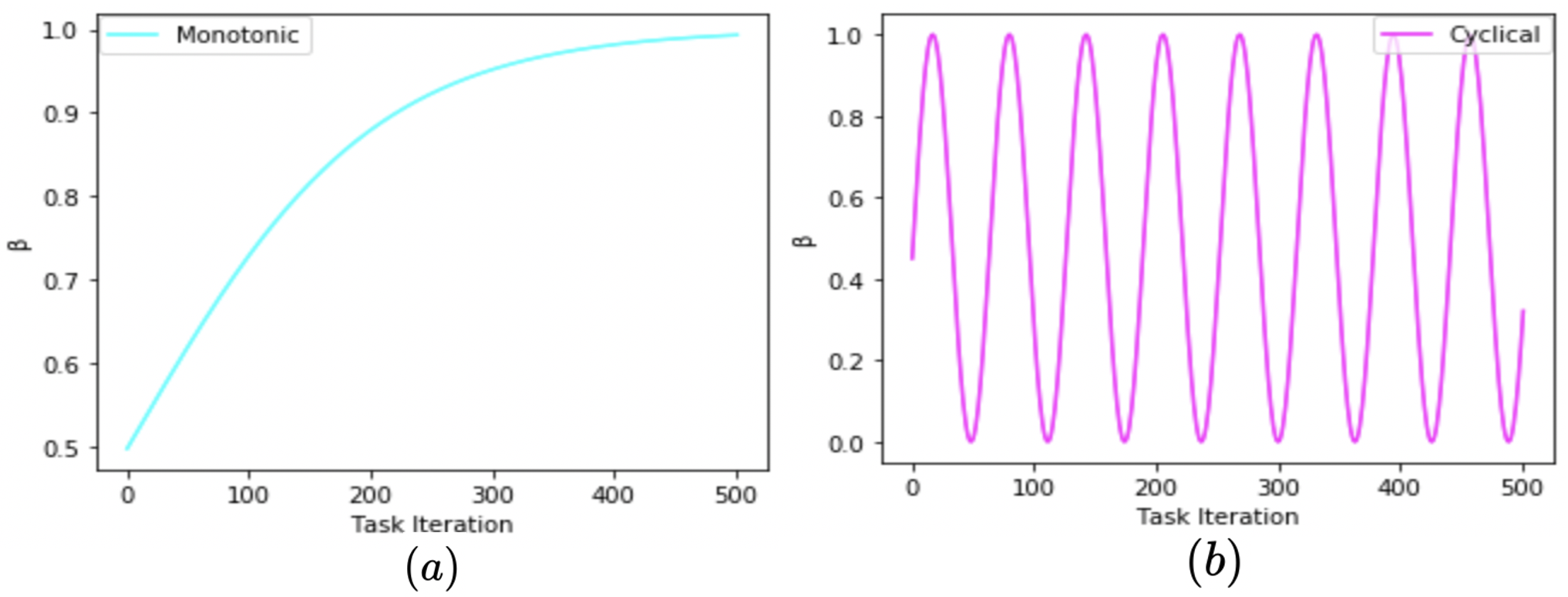}
\caption{Annealing $\beta$ with (a) monotonic schedule, (b) cyclical annealing schedule during multi-task learning (meta-training).}
\label{cyclical}
\end{figure}

Several attempts have been made to ameliorate the information preference problem in the VAE framework. Among them, the simplest solution is monotonic KL annealing, where the weight of the KL penalty term $\beta$ is scheduled to gradually increase during training ({\it monotonic schedule}) \cite{generatingsent} (see Figure \ref{cyclical} (a)). In the VAE framework, latent code $\phi$ learned earlier can be viewed as the initialization; such latent variables are much more informative than random and are thus ready for the decoder to use. Therefore, to mitigate the information preference problem, it is key to have meaningful $\phi$ at the beginning of training the decoder, so that $\phi$ can be utilized. Furthermore, \cite{cyclical} found that simply repeating the monotonic schedule multiple times ({\it cyclical annealing schedule}) enables high improvement in the above method (see Figure \ref{cyclical} (b)). 

We apply this type of regularization to the multi-task learning context. We basically consider the following objective that anneals the second term of the right-hand side in 
Eq.~\eqref{eq:neuralprocessesL} by the factor $\beta$ with $0 \leq \beta \leq 1$:
\begin{align*}
  & \displaystyle \hspace{0pt} \mathcal{L}_{\rm{MCA}} \!=\! 
  \mathcal{L}\left(\theta,\lambda\right) \!+\! \beta \ \mathbb{D}_{\rm{KL}}\left[ p(\phi| \tilde{x},\mathcal{D}_{S},\theta) || q_{\lambda}(\phi| \mathcal{D}_{S},\theta) \right].
\end{align*} 
We can see that by annealing the regularization term (second term), the effect of information preference property discussed in Section \ref{sec:InfPreProp} is mitigated. However, it is experimentally observed that just utilizing a fixed $\beta$ does not produce a good result. Thus, we gradually change the penalty term $\beta$ during the training. For that purpose, we decompose the objective into each task and control $\beta$ depending on each task as   
\begin{align}
\textstyle \mathcal{L}_{\rm{MCA}}   
= \sum^{T}_{t=1} \mathcal{L}_t\left(\theta,\lambda^{(t)}\right) + \beta^{(t)}\ \mathcal{R}^{(t)},
  \label{eq:MCA}
\end{align}
where 
$\mathcal{L}_t$ is the loss corresponding to a task $t$ and 
$\mathcal{R}^{(t)}$ is the regularization term for each training task.
During the training, we randomly sample task $t$ one after another and update the parameters where we apply different $\beta^{(t)}$ at each update. There could be several possibility of scheduling $\beta^{(t)}$ (Figure \ref{cyclical}), but we employ the cyclical annealing schedule during the training phase.
This realizes that each task can be trained with several different values of $\beta$ throughout the training, which yields avoiding the information preference problem.
Our experimental results show that this approach helps each task-specific learner to avoid falling into local minima. We call this approach as the meta cyclical annealing (MCA).

\subsection{Maximum mean discrepancy}
Unfortunately, the KL-term in the right-hand side in Eq. \eqref{eq:MCA} is difficult to compute. Therefore, we employ the maximum mean discrepancy (MMD) \cite{mmd} as a discrepancy measure between the distributions, which enables us to compute the corresponding term.
MMD is a framework to quantify the distance between two distributions by comparing all of their moments via a kernel technique. Letting $k(\cdot, \cdot)$ be any positive definite kernel, such as Gaussian kernel $k(\phi,\phi')=\exp(-{\frac{||\phi - \phi'||^2}{2\sigma^2}})$. 
MMD between $p$ and $q$ is defined as: $\mathbb{D}_{\rm{MMD}}\left[p(\phi)||q(\phi^{\prime})\right] = \mathbb{E}_{p(\phi),p(\phi^{\prime})}\left[ k(\phi, \phi^{\prime}) \right] + \mathbb{E}_{q(\phi),q(\phi^{\prime})}\left[ k(\phi, \phi^{\prime}) \right] - 2 \mathbb{E}_{p(\phi),q(\phi^{\prime})}\left[ k(\phi, \phi^{\prime}) \right]. $
It is known that if the kernel $k$ is {\it characteristic}, 
$\mathbb{D}_{\rm{MMD}}=0$ if and only if $p(\phi)=q_{\lambda}(\phi)$ \cite{muandet2017kernel}. A rough intuition of MMD is that difference of the moments of each distributions $p(\phi)$ and $q_{\lambda}(\phi)$ are measured through the (characteristic) kernel to know how different those distributions are. MMD can accomplish this efficiently via the kernel embedding trick.

We propose to employ MMD as the alternative divergence because MMD is easy to calculate and is stable against the support mismatch between the two distributions. To do so, instead of optimizing the objective introduced in Eq.\eqref{eq:MCA}, we minimize the following objective:
\begin{eqnarray}
  & \displaystyle \hspace{-9pt} \mathcal{L}_{\rm{MMD}}
  \!\!=\! 
  \mathcal{L}\left(\theta,\lambda\right) \!+\! \beta \!\  \mathbb{D}_{\rm{MMD}}\left[ p(\phi| \tilde{x},\mathcal{D}_{S},\theta) || q_{\lambda}(\phi| \mathcal{D}_{S},\theta) \right].~
\end{eqnarray}

\section{Experimental Results}
In this section, we experimentally show that 
the information preference problem of the posterior distribution actually occurs in non-regularized amortized inference and our proposal which aims to restrict the parameters of amortized distributions with MCA and MMD significantly improves the performance compared with existing methods. 

\subsection{Omniglot}
Omniglot \cite{omniglot} consists of 1623 characters from 50 different alphabets. Each of alphabets was hand drawn by 20 different people, thus 20 instances for each class (each character). We follow a pre-processing and training procedure by \cite{matchingnet} and \cite{versa}.

The training, validation and test sets consist of a random split of 1100, 100, and 423 characters, respectively. Each training iteration consists of the number of the mini batch that consists of random tasks extracted from the training set. During training, k-shot samples are used as training and remained 15 are used as test inputs. Evaluation after training is conducted on 600 randomly selected tasks from the test set. At the test phase, k-shot instances are utilized as test inputs which is unseen task for trained model. We use the Adam \cite{vae} optimizer with a constant learning rate of 0.0001 with 16 tasks per batch to train all models.

\subsection{mini-Imagenet}
The mini-ImageNet dataset consists of a subset of 100 classes from the ImageNet dataset \cite{imagenet} and contains 600 images for each class. Also 100 classes are divided into 64 train, 16 validation, and 20 test classes. This dataset is complex and difficult enough to evaluate few-shot classification tasks. Training proceeds in the same episodic manner as with Omniglot. 

\subsection{Effect of regularization}
\begin{figure}[]
\centering
\includegraphics[width=81mm]{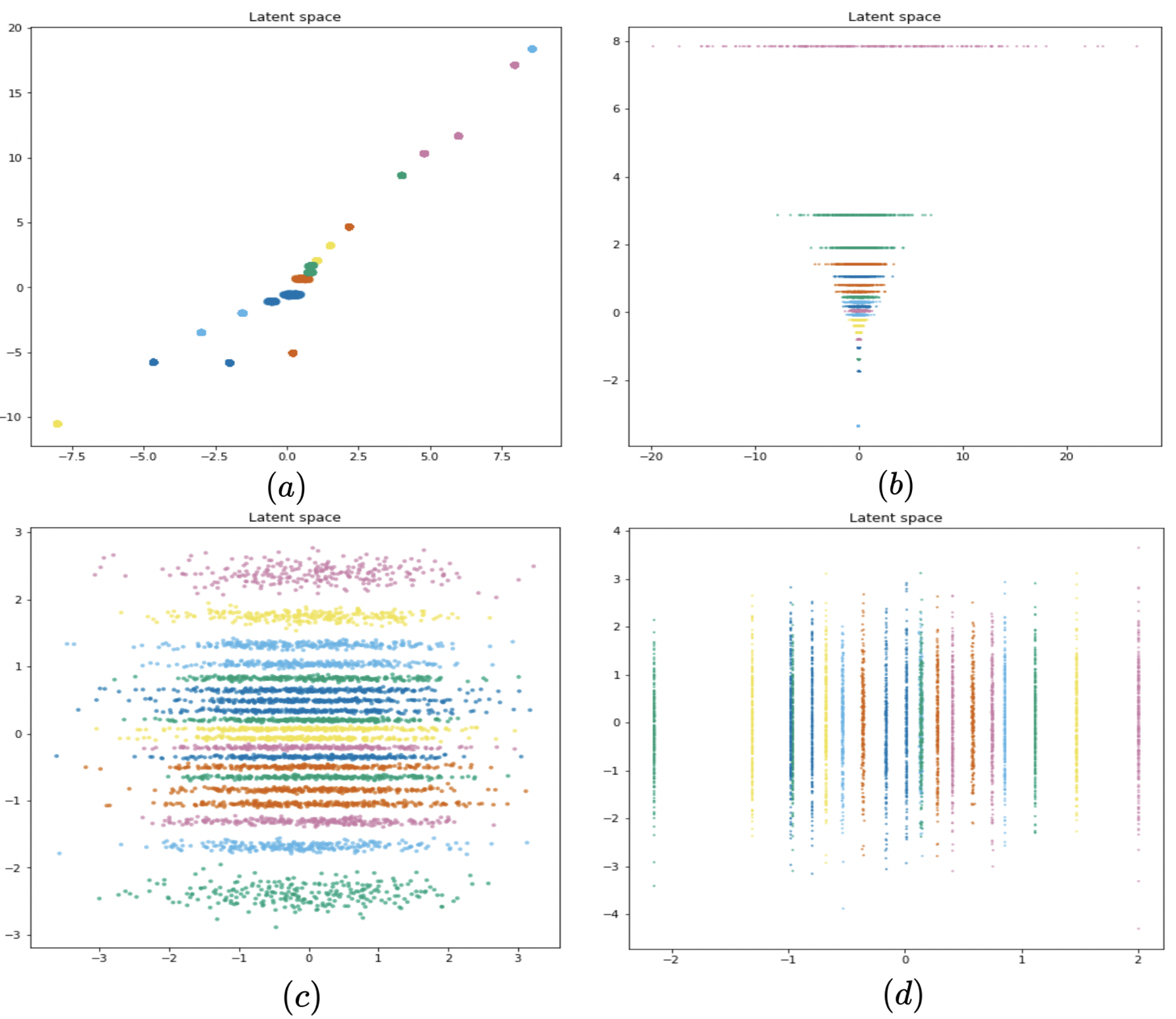}
\caption{Visualizing the learned latent code $\phi_{c} \in \mathbb{R}^{d}$ for $d = 2$. We examine 20-way 5-shot classiﬁcation in the Omniglot dataset. We randomly sample and fix fifty such tasks. Latent space with (a) the ordinary NPs, (b) NPs + MMD regularization without MCA (MMD+NPs), (c) NPs + meta cyclical annealing (MCA+NPs) and (d) NPs + meta cyclical annealing + MMD regularization (MCA+MMD+NPs).}
\label{latentspace}
\end{figure}

Here, we checked if our model estimates the latent code $\phi$ as expected via MCA and MMD. MCA and MMD regularizes the latent distribution close to standard Gaussian distribution to avoid information preference problem. Figure \ref{latentspace} shows the distributions of $\phi_c$. We can see that the distributions of MCA+NPs and MCA+MMD+NPs are well regulated and close to Gaussian distributions (see Figure \ref{latentspace} (c) and (d)). This is because of our MCA and MMD regularization. On the other hand, the distribution of NPs is far from Gaussians and the distribution of each class tends to degenerate to one-point (like a delta-distribution) which loses variation of $\phi_c$ resulting in worse posterior approximation (see Figure \ref{latentspace} (a)). This supports our expectation that the MMD-regularization effectively avoids the information preference problem. 

\subsection{Few-shot classification}
To compare with existing methods, we focus our method on standard few-shot classification tasks, 20-way classification for Omniglot and 5-way classification for mini-ImageNet. We do not evaluate 5-way classification for Omniglot because it is already set to more than 99\% with the existing methods, which is too high for comparing accuracy.

The results of Omniglot are shown in Table \ref{fewshot_omni}. Our proposal, MCA+NPs and MCA+MMD+NPs set good results. For 20-way 1-shot classification of Omniglot, our model achieves a new state-of-the-art result (99.81$\pm$0.14) which is significantly improved comparing with exiting methods. The result on mini-ImageNet is shown in Table \ref{fewshot_mini}. We see that, for mini-ImageNet, MCA+NPs achieves 77.37$\pm$1.67\% for 5-way 1-shot classification, MCA+MMD+NPs achieves 91.78$\pm$0.89\% for 5-way 5-shot classification. Both results are also new state-of-the-art. Furthermore, our experimental results demonstrate that our models surpass VERSA in terms of performance, which suggests that mitigating amortization error provides improvement.
\begin{table}
\hspace{-0.4cm}
\begin{minipage}[]{0.48 \textwidth}
\centering
\scalebox{0.9}{ 
\begin{tabular}{p{4.5cm}cccc}
\hline
&\multicolumn{2}{c}{\bf Omniglot}\\&\multicolumn{2}{c}{\bf 20-way ACCURACY(\%)}\\
\multicolumn{1}{l}{\bf METHODS}&\multicolumn{1}{c}{\bf1-shot}&\multicolumn{1}{c}{\bf5-shot} \\
\hline 
Matching Nets~\smallcite{matchingnet}	    &{93.8}&{98.5}\\
MAML~\smallcite{maml}
&{98,5$\pm$0.3}&{98.9$\pm$0.2}\\
Prototypical Nets~\smallcite{prototypical}	&{95.4}&{98.7}\\
Meta-SGD~\smallcite{meta-sgd}	            &{95.93$\pm$0.38}&{98.97$\pm$0.19}\\
Relation Net~\smallcite{relation-net}    	    &{97.6$\pm$0.1}&{99.1$\pm$0.1}\\
Reptile~\smallcite{reptile}    
&{89.43$\pm$0.14}&{97.12$\pm$0.32}\\
CNPs \smallcite{cnp}              
&{89.9}&{96.8}\\
VERSA~\smallcite{versa}              
&{97.66$\pm$0.29}&{98.77$\pm$0.18}\\
\hline
MCA+NPs                          &{\bf99.73$\pm$0.12}&{\bf99.9$\pm$0.21}\\
MCA+MMD+NPs                        &{\bf99.81$\pm$0.14}&{\bf99.9$\pm$0.17}\\
\hline
\end{tabular}
}
\caption{Accuracy comparison of few-shot classification.
The $\pm$ sign indicates the 95\% confidence interval. Bold text indicates the highest scores that overlap in their confidence intervals.
}
\label{fewshot_omni}
\end{minipage}
\vspace{-0.3cm}
\end{table}

\begin{table}[]
\hspace{-0.4cm}
\begin{minipage}[]{0.48 \textwidth}
\centering
\scalebox{0.9}{ 
\begin{tabular}{p{4.5cm}cccc}
\hline
&\multicolumn{2}{c}{\bf mini-ImageNet}\\&\multicolumn{2}{c}{\bf 5-way ACCURACY(\%)}\\
\multicolumn{1}{l}{\bf METHODS}&\multicolumn{1}{c}{\bf1-shot}&\multicolumn{1}{c}{\bf5-shot} \\
\hline 
Matching Nets~\smallcite{matchingnet}	    &{46.6}&{60}\\
MAML~\smallcite{maml}
&{48.7$\pm$1.84}&{63.11$\pm$0.92}\\
Prototypical Nets~\smallcite{prototypical}	&{46.61$\pm$0.78}&{65.77$\pm$0.70}\\
Meta-SGD~\smallcite{meta-sgd}	            &{50.47$\pm$1.87}&{64.03$\pm$0.94}\\
Relation Net~\smallcite{relation-net}    	    &{50.44$\pm$0.82}&{65.32$\pm$0.70}\\
Reptile~\smallcite{reptile}    
&{49.97$\pm$0.32}&{65.99$\pm$0.58}\\
VERSA~\smallcite{versa}              
&{53.40$\pm$1.82}&{67.37$\pm$0.86}\\
LEO~\smallcite{leo}  
&{61.76$\pm$0.08}&{77.59$\pm$0.12}\\
MetaOptNet-SVM~\smallcite{meta-opt}         
&{64.09$\pm$0.62}&{80.00$\pm$0.45}\\
ACC$+$Amphibian~\smallcite{amphibian}       
&{64.21$\pm$0.62}&{87.75$\pm$0.73}\\
\hline
MCA+NPs                          &{\bf77.37$\pm$1.67}&{\bf90.83$\pm$0.88}\\
MCA+MMD+NPs                        &{\bf76.67$\pm$1.46}&{\bf91.78$\pm$0.89}\\
\hline
\end{tabular}
}
\caption{Accuracy comparison of few-shot classification.
The $\pm$ sign indicates the 95\% confidence interval. Bold text indicates the highest scores that overlap in their confidence intervals.
}
\label{fewshot_mini}
\end{minipage}
\vspace{-0.3cm}
\end{table}

\section{Conclusions}
In this paper, we proposed the MCA+NPs and MCA+MMD+NPs models to improve amortized inference distribution with regularization techniques based on the latest few-shot learning framework, VERSA \cite{versa} and NPs \cite{np}. Through comparing methods on a common ground, our results show that the MCA+MMD+NPs model is comparable to state-of-the-art models under standard conditions, and the MCA+NPs model achieves comparable performance to recent state-of-the-art meta-learning algorithms on both the Omniglot and mini-ImageNet benchmark datasets. Additionally, our proposal seems to avoid the information preference problem by analysis.


\section*{Acknowledgments}
We would like to thank Naonori Ogasahara and Iwato~Amano for insightful discussions.~Taiji Suzuki was partially supported~by~JSPS Kakenhi (26280009, 15H05707 and~18H03201), and JST-CREST.

\bibliographystyle{named}
\bibliography{main}

\end{document}